# A Study on the Application of Artificial Intelligence in Ecological Design

Hengyue Zhao
School of Architecture
Carnegie Mellon University
Pittsburgh, PA, USA
hengyuez@andrew.cmu.edu

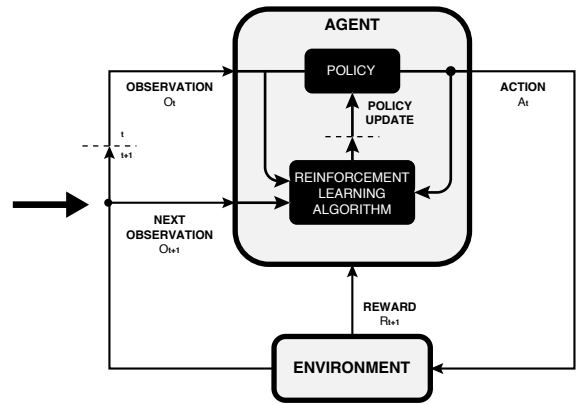

**Figure 1:** Three-stage roadmap: current limitations → RL agent → autonomous remediation. Left: today's system relies on manual sampling, fixed plant beds, isolated sensors, static LEDs, and labor-intensive upkeep. Center: a reinforcement-learning agent ingests real-time water-quality data, chooses control actions, and updates its policy from reward feedback. Right: the result is a self-optimising network—on-board heavy-metal sensors, adaptive hyacinth cartridges, solar relay rings, live LED feedback, and snap-fit modules for rapid swap-out.

## ABSTRACT

Can we acknowledge that our relationship with nature has evolved from human dominance to an intimate interconnectedness, recognizing that nature has genuinely attained a form of "personhood," and that artificial intelligence (AI) can facilitate this transformation, serving as a novel medium for human-nature connection? This article begins by examining the critical role of AI at the heart of the urgent ecological transformation currently underway, exploring the paradigm shift emerging from the intersection of AI and non-human life. The discussion progressively narrows its focus to how this innovative AI-nature paradigm manifests specifically within the fields of art and design, highlighting its distinctiveness from traditional artistic and design media.

The article seeks to explore how various artists and designers incorporate AI into ecological, microbiological, and geophysical creative practices. Through a comparative analysis of their creative strategies, it elaborates on the relationship between different applications of AI—such as data analysis, image recognition, and ecological restoration—and their unique artistic expressions, while also considering the extended value inherent in AI-driven art and design. However, the precise value of this emergent design paradigm remains subject to ongoing discourse.

Firstly, this new paradigm significantly innovates artistic methods and conceptual frameworks by integrating artificial intelligence and non-human life forms. Historically, design practices have primarily concentrated on the impacts of technology and nature upon human experience, rarely considering the creative potential unlocked when nature is technologically enhanced to inspire novel design paradig

Finally, drawing upon their own research, the authors propose innovative artistic approaches and initiate comprehensive discussions spanning theoretical foundations and practical applications. These discussions not only possess substantive academic significance but also offer fresh insights and perspectives conducive to further development within the field.
ms.

# 1 INTRODUCTION

## 1.1 Background of the Study

For centuries, the relationship between humans and nature has been characterized by a dynamic of domination and control. Humanity has successfully built a civilization that stands in contrast to natural systems, shaping and harnessing them for its own purposes. In this antagonistic relationship, civilization and nature have often been seen as distinct and opposing forces, with everything created by humans being defined as a product of civilization, while nature is often viewed as "the other." The way humans have controlled and manipulated nature has, to a large extent, led to ecological destruction.

To extend the boundaries of civilization, humans have devoted significant energy to infrastructure development, driven by motives such as imperial expansion, capital accumulation, and warfare. This process has been accompanied by rapid industrialization, urbanization, and commercial agriculture, all of which have exacerbated environmental degradation and increased pressure on natural resources and ecosystems. Today, the development of human society continues to be accompanied by a variety of serious ecological and environmental problems. If this trajectory continues, the planet will inevitably be destroyed by human actions. As a result, there is an urgent need to explore new pathways for ecological transformation to reshape the relationship between humans and nature.

The concept of ecological design emerged in the 1960s, coinciding with an increasing awareness of the environmental crisis and a growing focus on sustainable development. Ecological design gradually became one of the dominant paradigms in the design field. Its essence lies in humans' evolving understanding of the complexity of natural systems and the gradual abandonment of anthropocentric views. Over time, ecological design has become a design methodology that emphasizes environmental protection, resource efficiency, and social responsibility. Its core goal is to reduce the negative environmental impact of products and services throughout their life cycle, thus achieving the harmonious development of the economy, society, and environment.

Today, as we confront complex natural systems, it has become clear that humans can no longer fully comprehend the entirety of nature with our limited knowledge and abilities. This realization has prompted us to acknowledge that humans are merely a part of the world, and how we establish relationships with other entities in the world is a key issue. Therefore, ecological restoration is not only a remedial measure but also a way for humans to coexist with natural systems based on mutual benefit and compassion. This will involve the integration of science, art, social innovation, and design. In this context, the introduction of artificial intelligence (AI) as a new medium for human-nature relationships is explored, considering whether AI can become an important tool in establishing a new connection with nature and promoting ecological transformation.

## 1.2 Research Objectives, Methods, and Significance

This research explores the potential for ecological transformation through the integration of AI, using hybrid ecological practices powered by AI in plant-based projects. It synthesizes past ecological actions, design, art, and technology, particularly AI, to explore the direction of ecological transformation from multiple perspectives. By introducing new methods of AI involvement in ecological design, the research aims to rebuild the relationship between humans and nature, using bottom-up approaches to cultivate new lifestyles and ultimately alleviate the current ecological crisis.

Through methods such as literature review, case analysis, inductive reasoning, and comparative research, this study analyzes the challenges faced in ecological environmental protection in China.

It explores the necessity of AI in ecological design across five areas: accuracy, efficiency, intelligence, and sustainability. The paper proposes three intervention paths—intelligent early warning, intelligent supervision, and intelligent maintenance—aiming to enhance the protection of ecological environments.

This research also discusses how AI can foster a closer connection between humans and nature and proposes design interventions for ecological protection, providing new models for environmental conservation.

## 1.3 Research Status at Home and Abroad

In recent decades, many artists and designers worldwide have begun to engage with the ecological domain. For example, Buckminster Fuller's 1960 proposal for the Manhattan Dome aimed to regulate the weather and reduce air pollution. Joseph Beuys, one of the early artists involved in ecological action, carried out the Elbe River cleanup project in Hamburg in 1962. Agnes Dénes, in 1982, planted and harvested a wheat field on a landfill site in downtown Manhattan, worth $4.5 billion, using the field as a symbol of food, energy, trade, and economic issues, while reflecting on mismanagement, waste, global hunger, and ecological problems. Mel Chin's "Revival Field" explored the use of hyperaccumulating plants for soil remediation. These plants absorb toxic heavy metals like cadmium, zinc, and nickel from the soil, which are then harvested and recycled. This process, known as "phytoremediation," involved significant cross-disciplinary collaboration.

However, these artistic works still have limited perspectives and lack a systemic approach to addressing the root causes of ecological crises. They also fail to explore the role of technology in addressing ecological crises.

Currently, research on AI's involvement in ecological design and its role in promoting ecological transformation is still limited, with attempts being made in areas such as interior design, architectural design, and rural ecological design. In this context, John Thackara has proposed new ideas in the field of AI-powered ecological transformation. He argues that the relationship between humans and nature should shift from one of control to one of kinship, establishing ecological restoration that transcends human limitations. Thackara emphasizes the importance of natural connection and how farm visits can promote this connection, introducing the concept of "citizen ecology" and the TreesAI platform, which helps urban managers improve the planting and maintenance of urban forests to address environmental issues. Overall, he stresses the importance of strengthening the reciprocal and sustainable relationship between humans and nature.

Meanwhile, domestic scholars have also proposed valuable insights in their respective research fields. For instance, Liang Jianian and Yang Yuxin believe that AI can significantly enhance the ecological design of interior environments, while Cao Shengsheng advocates for AI to improve rural ecological design, thereby promoting sustainable rural development. Guan Zenglun has highlighted AI's role in advancing green smart buildings, which have vast potential. Similarly, Liu Zhixiong and Li Yanfei argue that the development of the digital economy can empower ecological environmental protection. Researchers such as Fei Yanxiao, Wu Junxing, Le Weiqing, and Linghu Xingbing have discussed the role of AI and 5G technologies in smart environmental protection and ecological management, all concluding that AI's integration can enhance ecological conservation efforts.

In summary, research on AI's integration into ecological design is still in its early stages, with limited theoretical and practical studies. The relevant literature and artistic practices provide important references for this paper.

## 2 EXPLORING THE POTENTIAL OF INTEGRATING AI AND ECOLOGICAL DESIGN TO PROMOTE ECOLOGICAL TRANSFORMATION

### 2.1 Artificial Intelligence and Ecological Design

With rapid socio-economic growth and urbanization in China, improvements in living standards have been accompanied by various ecological and environmental challenges, including air and water pollution, grassland degradation, deforestation, and biodiversity loss. Traditional ecological design approaches often exhibit critical limitations, such as reliance on designers' subjective experiences, insufficient scientific methods, and inadequate tools, resulting in designs that lack accuracy and reliability. Additionally, traditional approaches rarely incorporate comprehensive data analysis, leading to suboptimal outcomes due to a failure to fully consider multiple influencing factors. Furthermore, traditional ecological design typically focuses on current conditions without adequately accounting for future changes and uncertainties, lacking optimization capabilities and the capacity for rapid identification and correction of design flaws through simulation and modeling. Consequently, current technological approaches and intervention methods are insufficient to resolve environmental pollution issues, address the limitations inherent in traditional ecological design, and enhance ecological design systems. Artificial Intelligence (AI), first introduced in 1956 by computer scientist John McCarthy, is a branch of computer science concerned with theories, methods, techniques, and applications that simulate, extend, and augment human intelligence. AI primarily involves storing knowledge and enabling programs to achieve predefined goals through computational methods. As AI technology advances, its application in ecological design is becoming increasingly prevalent. The growing sophistication of AI technology continuously drives innovation in design, addressing key shortcomings of traditional ecological design, such as accuracy and timeliness. For example, machine learning and data mining techniques enable the discovery and analysis of environmental and resource trends based on extensive datasets and models, providing designers with more comprehensive and precise data support. AI can also employ modeling and simulation systems to perform scenario analyses and predictions, thereby optimizing design solutions and enhancing overall efficiency. Moreover, the integration of AI broadens design thinking and introduces fresh perspectives into contemporary ecological design practices, paving new pathways toward creating pollution-free, healthy living environments and sustainable habitats.

### 2.2 The Necessity of Integrating AI into Ecological Design

The history of modern humanity is characterized by a growing separation from nature. Our excessive pursuit of development has often led us to neglect ecological concerns. Human societies typically prioritize economic growth, overlooking the foundational role that ecosystems play in supporting economies. This oversight has significantly harmed the environment that sustains our lives.

As ecological crises and environmental issues grow increasingly complex, human capabilities alone are insufficient to address these challenges. Integrating artificial intelligence (AI), as another complex system, into ecological design offers a viable solution:

1. AI-driven ecological design emphasizes the purpose and outcomes of design, beyond mere aesthetics and form.
2. AI provides intelligent analysis and optimization of environmental and resource data, significantly enhancing design efficiency and accuracy.
3. AI-driven design leverages extensive datasets and models through machine learning and data mining techniques to detect and analyze environmental trends, thereby offering designers more comprehensive and accurate data support.
4. AI-driven ecological design considers entire ecosystems, taking into account diverse factors to achieve sustainable design outcomes that protect the environment and human health.
5. AI-driven design facilitates modeling and simulation, enabling scenario-based predictions and rapid identification and correction of design flaws, thus optimizing design processes.

Therefore, developing AI-integrated modern ecological design is an inevitable trend. With AI assistance, harmonizing the relationship between humans and nature holds significant potential for addressing environmental challenges such as air and water pollution and climate change.

## 2.3 Pathways for Integrating AI into Ecological Design

With the increasing prominence and widespread adoption of sustainable development concepts, enhancing ecological environments has gained heightened attention. Particularly with rapid economic growth, harmonizing ecological conservation and high-quality development has emerged as a critical issue today.

In the AI era, adopting advanced technologies such as satellite remote sensing, drone inspections, real-time monitoring, and big data analysis enables "non-contact" and "remote" management approaches. These technologies significantly enhance environmental management efficiency, reduce pollution, prevent environmental crises, and substantially improve ecological outcomes. Integrating AI into ecological design to promote environmental protection can be approached through three primary pathways: intelligent early warning, intelligent supervision, and intelligent maintenance.

Firstly, high-density monitoring stations for fixed pollution sources and key regions utilize AI for deep learning data analysis, enabling rapid response and intelligent early-warning systems, with real-time alerts disseminated through data platforms and mobile applications.

Secondly, intelligent supervision employs sensors for real-time monitoring of target areas. The sensor data are transmitted to data centers for comprehensive analysis, allowing precise and scientifically grounded environmental management solutions that surpass traditional manual supervision methods.

Finally, intelligent maintenance introduces the Internet of Things (IoT) and 5G technologies into environmental education and advocacy, fostering greater public awareness and participation in ecological protection. Additionally, applying AI technologies in daily production and life conserves human resources and reduces water and electricity consumption, effectively maintaining ecological balance.

Currently, integrating AI into ecological design in China is still at a preliminary exploration stage. The strategic implementation of intelligent early warning, intelligent supervision, and intelligent maintenance can effectively enhance ecological protection efforts.

## 2.4 Case Studies: AI-Enabled Ecological Design Projects

Currently, integrating AI into ecological design in China is still at a preliminary exploration stage. The strategic implementation of intelligent early warning, intelligent supervision, and intelligent maintenance can effectively enhance ecological protection efforts.

*2.4.1 **Restor: Scaling Nature-Based Restoration***

Restor is an AI-driven global platform dedicated to accelerating nature conservation and restoration for the benefit of people, biodiversity, and climate. By connecting practitioners with scientific data, monitoring tools, funding sources, and one another, Restor seeks to amplify the impact, scale, and sustainability of restoration efforts. Users can delineate an area of interest on an interactive map and immediately receive insights into local biodiversity, current and potential soil carbon, land-cover patterns, soil pH, annual precipitation, and other variables. An integrated dashboard leverages high-resolution satellite imagery and AI-based analytics to track site changes over time and visualize progress via standardized ecological indicators. By encouraging users to publish their projects, the platform increases transparency and fosters knowledge exchange, thereby accelerating global restoration actions. In Restor, AI functions as an engine for environmental analysis, continuous monitoring, and community mobilization, markedly increasing the efficiency and efficacy of restoration work.

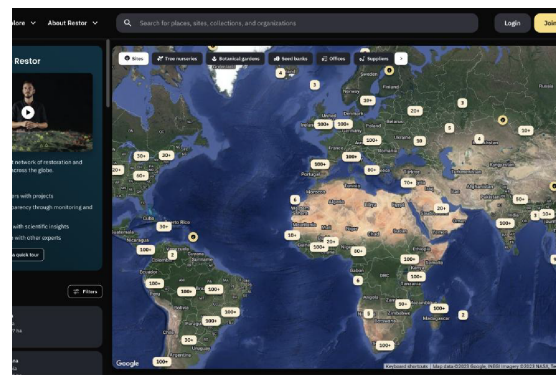

**Figure 2: The Restor platform visualizes global ecological restoration sites, allowing users to explore biodiversity, carbon storage, and land data through AI-powered satellite analysis and interactive map filters.**

*2.4.2 BirdNET: Acoustic Monitoring for Avian Conservation*
BirdNET is a research platform that applies machine-learning techniques to the large-scale detection and classification of bird vocalizations. Developed jointly by the K. Lisa Yang Center for Conservation Bioacoustics at the Cornell Lab of Ornithology and the Chair of Media Informatics at Chemnitz University of Technology, the project provides innovative tools for ornithologists, conservationists, and citizen scientists. BirdNET supports a wide range of hardware and operating systems, including Arduino microcontrollers, Raspberry Pi devices, smartphones, web browsers, workstations, and cloud services, enabling flexible deployment in diverse field conditions. The current model recognizes approximately 3,000 of the world's most common bird species, with ongoing expansion planned. By combining AI-based acoustic analysis with scalable field hardware, BirdNET has become a critical asset for real-time avian monitoring and conservation, demonstrating AI's pivotal role in species detection, data management, and decision support.

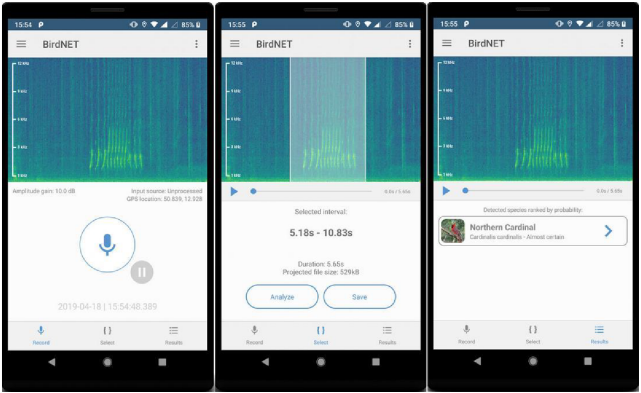

**Figure 3: Screenshots of the BirdNET mobile interface, which enables users to record bird sounds, visualize spectrograms, and receive AI-based species identification results in real time.**

*2.4.3 AudioMoth: Low-Cost, AI-Assisted Bioacoustic Recording*
AudioMoth is a low-cost, compact device designed to facilitate the comprehensive and accurate recording of wildlife sounds for biodiversity and ecosystem monitoring. Its affordability permits large-scale deployments, while its small form factor and low power consumption make it suitable for remote and long-term studies. An adjustable-gain analog pre-amplifier and configurable USB interface allow the device to support a variety of recording tasks. Compared with traditional wildlife-monitoring methods, Audio-Moth excels at capturing small or elusive species such as bats and woodpeckers, and its full-spectrum recording capability extends to insects, amphibians, and mammals. AI-enabled post-processing pipelines perform automated sound classification and data analysis, providing researchers with rich, high-quality datasets that reveal interspecies interactions and ecosystem dynamics. Consequently, AI within the AudioMoth workflow significantly enhances both the precision and scope of wildlife monitoring.

Taken together, these case studies demonstrate AI's multifaceted contributions to ecological design—spanning environmental assessment, species identification, and participatory restoration. By automating labor-intensive tasks and delivering granular ecological insights, AI technologies are helping designers, scientists, and local communities to co-create more resilient and sustainable ecosystems in China and beyond.

## 3    "HEAVY METAL LOCK" ARTISTIC DESIGN PRACTICE: EMPOWERING ECOLOGICAL AND CREATIVE FORMS THROUGH NEW TECHNOLOGIES

### 3.1 Design as Research: Ecological Blind Spots Created by Electronic Devices

*3.1.1 Heavy Metal Pollution Caused by Electronic Waste*
In recent decades, the rapid development of digital devices has brought immense convenience to our lives, but it has also led to a series of environmental issues. The extensive pollution footprint left by large-scale digital industries and logistics systems, such as electronic waste and chemical emissions, poses serious threats to soil, water, and air quality. These problems cannot be ignored as they directly impact our health and ecological balance. On the one hand, the widespread use of digital technology has resulted in the accumulation of electronic waste and the difficulty of managing it.

Research shows that electronic waste contains over 700 substances, including plastics, gold, silver, and copper, which can be recycled. However, there are also numerous harmful heavy metals such as lead, mercury, chromium, cadmium, and nickel, along with persistent organic pollutants (POPs) like PVC, PCBs, and PBDEs. If these substances cannot be effectively recycled, they will cause significant harm to both our ecological systems and human health, particularly since the heavy metals in electronic waste are mostly non-degradable toxic substances. These substances cannot be broken down by biological processes and lack natural purification capabilities. Once they enter the environment, it becomes extremely difficult to remove them.

According to data, global electronic waste production was approximately 48.9 million tons in 2012, with projections for 2017 reaching 65.4 million tons. Due to economic development, China's annual electronic waste growth rate is between 3% and 5%, including old equipment such as phones and computers. Due to cheap labor and lagging environmental regulations, China has become the dumping ground for electronic waste from developed countries, with 80% of the U.S. electronic waste sent to Asia, 90% of which ends up in China. In China, towns like Guiyu in Guangdong Province and Taizhou in Zhejiang Province are the focal points of electronic waste disposal, which has spread to other areas. Sampling analysis of the Lianjiang River in Guiyu revealed serious heavy metal contamination, with 18 metals exceeding safe limits, indicating severe pollution.

Waterborne heavy metal pollution is another major issue today. Heavy metals in water can be categorized into two sources: natural and anthropogenic. Naturally occurring heavy metals are released from rocks and soils when groundwater or surface water flows through metal-rich areas, and can also be carried in rainfall runoff. Anthropogenic pollution arises from human activities such as industrial production wastewater, mining wastewater, and the discharge of pollutants from transportation and energy industries. Industrial wastewater includes heavy metals from industries such as smelting, electroplating, chemicals, tanning, medicine, coatings, and pesticides. The improper discharge and treatment of industrial wastewater is one of the main causes of water pollution.

*3.1.2. Analysis of the Causes of Heavy Metal Pollution*
The main causes of heavy metal pollution due to electronic waste are threefold. First, the raw materials used in electronic devices contain significant amounts of heavy metals, such as copper, lead, and nickel, which are essential in processes like PCB manufacturing. For instance, in the electronic industry's PCB manufacturing process, multiple stages such as pre-treatment, drilling, and exposure require the use of heavy metals.

Second, the manufacturing of electronic devices often necessitates the addition of heavy metals. For example, the assembly of various digital devices involves adding different metals, such as lead-free solder and soft soldering materials, which are required to ensure the appropriate melting point, wetting, flexibility, and filling capacity of the solder.

Finally, improper disposal of discarded electronic devices contributes to heavy metal pollution. These devices are either discarded improperly or treated using outdated recycling methods, such as pyrometallurgical and hydrometallurgical processes, which directly release large amounts of heavy metals into the environment. For example, during hydrometallurgical processes, acid washing or acid baths produce waste acid that contains high concentrations of copper ions, which are often indiscriminately dumped into rivers or soil, causing secondary pollution.

### *3.1.3. Ecological Blind Spots*

I introduce the concept of "ecological blind spots," which stems from our overreliance on technology, leading to reduced attention to natural ecology. For example, the massive digital infrastructure, logistics, economic systems, and trade networks behind devices like smartphones contribute to ecological damage with every use. Due to our focus on the digital world, we often fail to notice these potential, silently existing pollutants, which constitutes an ecological blind spot.

The environmental impact of games like World of Warcraft and Fortnite is another blind spot; these games consume an average of 500 watts of electricity per hour, generating significant amounts of $CO_2$. We also encounter blind spots in areas such as overexploitation of resources and economic development at the cost of the environment. For instance, the production of a smartphone generates approximately 60 kilograms of $CO_2$.

The effects of these blind spots are deeply harmful to our ecology. Why are we unable to perceive them? Because most of our attention is focused on managing digital information systems, leaving little room to pay attention to natural ecosystems, let alone finding solutions. In this context, AI may offer an effective new approach to addressing these crises.

## 3.2 Design as a System: Next-Generation Waterborne Heavy Metal Treatment System

Based on the research above, the designer's work Heavy Metal Lock focuses on empowering ecological design through new technologies, offering fresh perspectives on both our ecological design systems and governance methods. This chapter will explore how technology can empower ecological design, focusing on the form and content of the creation, and how it brings about new changes.

In recent years, global attention to heavy metal pollution in water has grown significantly. Traditional methods of treating waterborne heavy metals are often costly and unsustainable. Several physical, chemical, and biological technologies exist for restoring soils contaminated by heavy metals; however, these methods have limitations. For example, these processes require substantial effort and incur high costs, and traditional remediation methods can cause further harm to the soil, leading to irreversible changes in the physical and chemical properties of the soil. Therefore, it is crucial to find an economically efficient solution for heavy metal remediation, and plant-based remediation presents a viable option. Building on plant-based remediation, innovative systems have been designed to address waterborne heavy metal pollution.

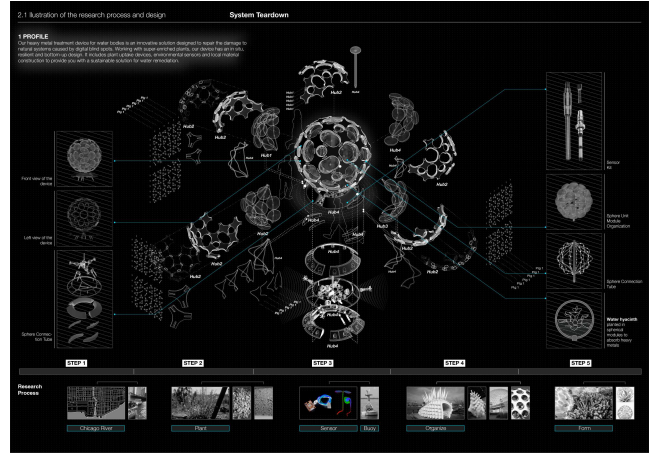

**Figure 4: An overview of the modular water remediation system, showing the five-step design process integrating plants, sensors, and bio-modular spheres for in-situ heavy metal treatment.**

For this purpose, the designer has created a new waterborne heavy metal treatment system, drawing inspiration from botany and collaborating with hyperaccumulating plants to adopt a more localized, flexible, and bottom-up design approach. First, we leverage knowledge from botany to study the capacity of hyperaccumulating plants in absorbing, transforming, and immobilizing heavy metals. By harnessing the biological processes of these plants, we can use them to restore water bodies polluted by heavy metals. This method involves six main types: phytoremediation, phytodegradation, rhizoremediation, phytoextraction, phytovolatilization, and phytostabilization. Typically, plants absorb heavy metals from the water, store them as biologically available metals and nutrients, and these are retained within the plant system. Factors such as water pH, organic matter content, root exudates, microbial biomass, and competitive cations influence the efficiency of heavy metal uptake. The plant extraction mechanism includes five key steps: heavy metal enrichment in the rhizosphere, absorption by plant roots, transport to the aerial parts of the plant, chelation within plant tissues, and heavy metal tolerance. The tolerance of plants to heavy metals is a prerequisite for successful remediation. The higher the tolerance, the more heavy metals can be accumulated with less negative impact on plant health. Beyond controlling plant tolerance mechanisms, annual biomass yield (e.g., stem and branch weight) and net metal harvest per year are critical for assessing the extraction capacity of plants.

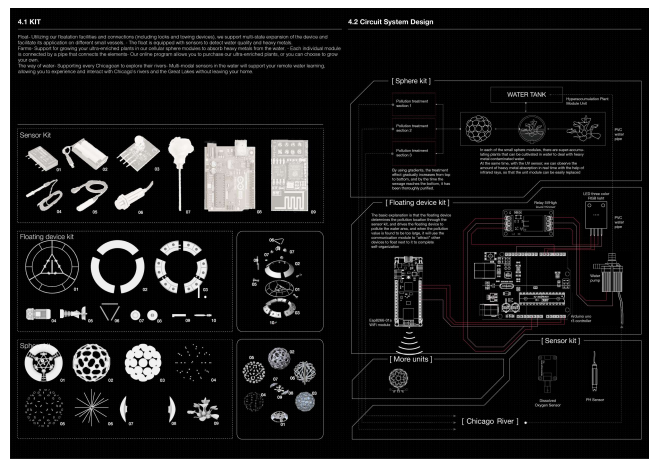

**Figure 5: System kits and circuit layout for modular water remediation.**

Secondly, we select plant species suitable for the local environment, utilizing their absorption capacity to remove heavy metal pollutants from water bodies. This natural bioremediation method is not only effective but also has low cost and minimal environmental impact.

Thirdly, we adopt a localized design strategy that involves the active participation of local communities, ensuring the adaptability and sustainability of the solution. By collaborating with local expert systems, we can more accurately pinpoint the sources and spread of heavy metal pollution in water bodies. The study of plant expert systems, which began in the 1960s, has evolved from simple plant databases to complex decision support systems used for crop pest diagnosis. These systems have been applied in areas such as crop management and ecological conservation, with notable examples of successful plant expert systems in the U.S., Japan, and the Netherlands.

Finally, using locally available materials, such as plant absorption units, supporting bases, and water environment sensors, we can design a water treatment system tailored to local conditions and needs. To improve the efficiency and convenience of setting up the system, we have developed a user interface that allows users to access design blueprints, purchase seed packages for planting, and assemble the system. Users can also upload data on local water pollution conditions and, when the crisis worsens, initiate a collaboration call for others to join in solving the heavy metal pollution problem.

In the design of Heavy Metal Lock, AI plays a pivotal role by enabling intelligent and automated water treatment systems. Here are some key AI attributes in this process:

1. Prediction and Pattern Recognition: AI models predict and identify heavy metal pollution in water bodies in real-time, providing accurate early warnings, making remediation efforts more targeted and timely.
2. Adaptability: AI systems can adjust parameters like the angle, height, and absorption area of the plant absorption devices in real-time, ensuring optimal remediation results.
3. Data Analysis and Decision Support: AI systems automatically collect and analyze environmental data, providing decision support for optimizing plant-based remediation strategies.
4. Learning and Self-Optimization: The system continuously optimizes its predictive models and decision rules through ongoing data collection, improving the efficiency of remediation over time.
5. Interactivity and Community Participation: The system can automatically generate remediation plans based on user-uploaded pollution data, assist with seed package purchases, and guide the assembly of devices. It can also help initiate community engagement by inviting like-minded locals to participate in the restoration efforts.

These attributes not only enhance the efficiency of remediation but also lower costs and foster community involvement, offering strong technological support for realizing the design vision.

## 4   CONCLUSION

Humanity's millennia-long pursuit of taming and controlling nature has culminated in severe ecological degradation and resource depletion. Silent Spring sounded a collective alarm, reminding us that the human–nature relationship must be fundamentally re-conceived. Ecological design offers new lenses and tools for this task, yet today's intertwined natural systems demand a globally informed, bottom-up perspective. Against this backdrop, we interrogate whether artificial intelligence (AI) can serve as a catalytic instrument for achieving harmonious coexistence—a question that grounds and motivates this study.

Focusing on China's pressing crisis of heavy-metal contamination in freshwater bodies, we observe clear limitations in conventional ecological-design workflows: heavy reliance on designers' experience, fragmented data, and subjective judgement often leave critical "blind spots." Rapid advances in AI now furnish richer environmental data, predictive modelling, and optimisation capabilities that can systematically close these gaps. AI's capacity to analyse, forecast, and iteratively refine complex eco-systems positions it as an indispensable partner in the pursuit of sustainable solutions.

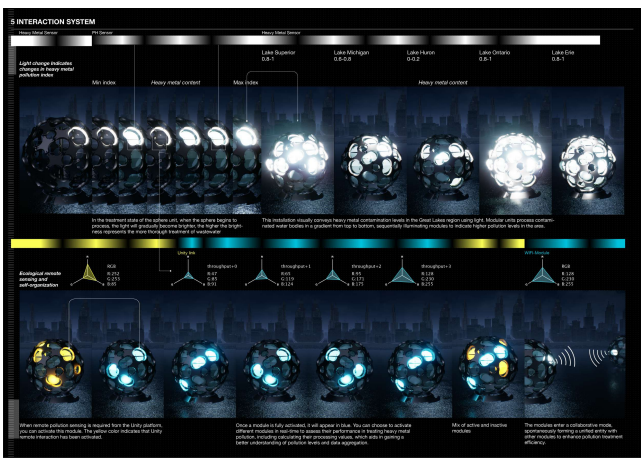

**Figure 6: Visualization of the interaction system, where modular sphere units use light intensity and color to indicate heavy metal contamination levels. The system integrates real-time sensing, remote activation, and collaborative self-organization to enhance water pollution monitoring and treatment.**

To illustrate these claims, we articulated three AI-enabled design pathways—intelligent early warning, intelligent monitoring, and intelligent maintenance—and demonstrated how each pathway measurably enhances environmental-protection practices: improving management efficiency, reducing pollution, and ultimately restoring ecological balance.

Re-examining decades of digital hyper-consumption, we recognise the hidden ecological costs of electronic waste, soaring energy demand, electromagnetic radiation, and associated carbon emissions. By reframing AI not as yet another extractive technology but as a mediating medium that cultivates intimacy—rather than dominion—between humans and nature, we outline a paradigm shift from control to care. Concretely, we proposed and prototyped an AI-augmented device for in-situ remediation of heavy-metal pollutants in water, thereby mitigating damage caused by digital hardware itself.

Our work aspires to move public discourse from "design for consumption" toward "design for crisis." By exposing the latent environmental footprint of everyday devices and demonstrating actionable remediation strategies, we aim to heighten ecological awareness and encourage collective responsibility for mitigating current—and future—environmental crises.

## ACKNOWLEDGEMENTS

I would like to express my sincere gratitude to my advisor, Professor Ziyuan Wang, whose professional guidance and insightful suggestions were instrumental in the completion of this research. His support and mentorship, especially during moments of uncertainty, were deeply appreciated and invaluable.
I would also like to thank my course instructors and fellow classmates for their encouragement and generous support throughout my academic journey.
Finally, I am truly grateful to my family and friends. Their unwavering encouragement gave me the strength to persevere and complete this work. Their support will always remain close to my heart.